\documentclass{Interspeech2024}

\interspeechcameraready

\title{Exploring Self-Supervised Multi-view Contrastive Learning for Speech Emotion Recognition with Limited Annotations}

\name[]{Bulat}{Khaertdinov$^*$} 
\name[]{Pedro}{Jeuris$^*$}
\name[]{Annanda}{Sousa}
\name[]{Enrique}{Hortal}

\address{
  Maastricht University, the Netherlands}
\email{\{b.khaertdinov, pedro.jeuris, annanda.sousa, enrique.hortal\}@maastrichtuniversity.nl}

\keywords{speech emotion recognition, self-supervised learning, contrastive learning, sparse annotations}

\usepackage{multirow}
\usepackage{amssymb}
\usepackage{pifont}
\newcommand{\cmark}{\ding{51}}%
\newcommand{\xmark}{\ding{55}}%
\usepackage{tablefootnote}
\usepackage{graphicx}
\usepackage{subcaption}

\begin{document}

\maketitle
\def\thefootnote{*}\footnotetext{Authors contributed equally to this work}\def\thefootnote{\arabic{footnote}}

\begin{abstract}
    Recent advancements in Deep and Self-Supervised Learning (SSL) have led to substantial improvements in Speech Emotion Recognition (SER) performance, reaching unprecedented levels. However, obtaining sufficient amounts of accurately labeled data for training or fine-tuning the models remains a costly and challenging task. In this paper, we propose a multi-view SSL pre-training technique that can be applied to various representations of speech, including the ones generated by large speech models, to improve SER performance in scenarios where annotations are limited. Our experiments, based on wav2vec 2.0, spectral and paralinguistic features, demonstrate that the proposed framework boosts the SER performance, by up to 10\% in Unweighted Average Recall, in settings with extremely sparse data annotations.
    
\end{abstract}

\section{Introduction}
Emotion recognition is a crucial task of Affective Computing that gained a significant amount of research attention in the last decades. Speech serves as a key marker for effective emotion recognition, encompassing diverse acoustic, prosodic, and other voice-related information and accounting for inter-speaker differences \cite{ghaleb_dissertation}. During the last ten years, Speech Emotion Recognition (SER) algorithms have been significantly improved due to the rapid development of Deep Learning architectures. The earlier methods of the decade were based on the end-to-end supervised Deep Learning models exploiting prosodic or spectral features \cite{eyben2015geneva, deng2014autoencoder, zheng2015experimental}, or raw audio waveforms \cite{trigeorgis2016adieu}. In the last couple of years, the research focus has been shifting towards exploiting Transformer-based large speech models, such as wav2vec 2.0 \cite{baevski2020wav2vec}, WavLM \cite{chen2022wavlm}, and HuBERT \cite{hsu2021hubert}, pre-trained via Self-Supervised Learning frameworks \cite{pepino21_interspeech, wang2021fine, morais2022speech, zaiem2023speech}. 

One of the key challenges always associated with emotion recognition is collecting data with trustworthy annotations \cite{schuller2021review}. Furthermore, emotion recognition systems could be deployed in various scenarios requiring data collection in natural settings and utilizing specific emotional models that are not covered in open access data \cite{rudovic2019personalized, singh2021multimodal, mousavi2023emotion}. In this case, acquiring a dataset even with hundreds of samples containing effectively elicited emotions and accurate annotations is an extremely challenging and time-consuming process \cite{akccay2020speech}. Deep learning models trained from scratch typically require large amounts of accurately annotated data to achieve satisfactory performance, whereas large pre-trained models can be fine-tuned with less, but still significant, amounts of annotated data.

In this paper, motivated by these challenges, we introduce multi-view contrastive SSL pre-training that can be applied on top of various audio features (views), including paralinguistic cues, spectral representations, and features extracted by large speech models pre-trained on ASR datasets. The contributions of this work can be summarized as follows:
\begin{itemize}
    \item The introduced framework, denoted as Pairwise-CL, aims to pre-train encoders on multiple speech views for further fine-tuning with sparsely annotated data. Pre-training is based on contrastive SSL loss computed between representations of speech views in a pairwise fashion. Specifically, the encoders from the selected views aim to align representations of each utterance in the projected latent space.
    \item The proposed framework can be adapted to any combination and number of views. The experiments in this paper were conducted on three types of views, namely wav2vec 2.0 features \cite{baevski2020wav2vec}, mel spectrograms and eGeMAPS-88 \cite{eyben2015geneva}. 
    \item We analyze the representations learnt from each view and quantify their alignment using projection-weighted Canonical Correlation Analysis (PWCCA) \cite{morcos2018insights}.
\end{itemize}

\section{Methodology}
\subsection{Pairwise-CL: Multi-view Contrastive Learning}
\label{sec:multi-view}

In the last years, contrastive Self-Supervised Learning has shown promising results in multi-modal and multi-view pre-training in different domains \cite{tian2020contrastive, brinzea2022contrastive, xu2022multi, tu2023context, girdhar2023imagebind}. The main idea lies in maximizing similarities between different representations of the same instance in a projected latent space while contrasting them to other instances. The pre-training strategy introduced in this paper is inspired by Contrastive Multiview Coding (CMC) \cite{tian2020contrastive} suggested for multi-view image representation learning. Namely, we propose using normalized temperature-scaled cross-entropy loss (NT-Xent) \cite{chen2020simple} in between pairs of view-level representations corresponding to different audio features. We denote the proposed framework as Pairwise-CL.

Formally, assume there is a mini-batch of size $N$ with features from $K$ views $\{f_1(\boldsymbol{x}_l^1), f_2(\boldsymbol{x}_l^2), \dots, f_K(\boldsymbol{x}_l^K)\}_{l = 1}^N$ where $f_i(\cdot): \mathcal{X}_i \rightarrow \Phi_i \subset \mathbb{R}^{d_i}$ is a view-level encoder mapping inputs $\boldsymbol{x}_l^i \in \mathcal{X}_i$ from $i$-th view to a vector of size $d_i$. The view-level representation dimensionality $d_i$ is based on the encoder architecture processing the view. Then, the features from each view are mapped to the space where contrastive loss is computed using separate projection networks $g_i: \Phi_i \rightarrow \Lambda \subset \mathbb{R}^D$, i.e. $\boldsymbol{z}_l^i = g_i(f_i(\boldsymbol{x}_l^i))$. Thus, the set of projected representations can be written as $\{\boldsymbol{z}_l^1, \boldsymbol{z}_l^2, \dots, \boldsymbol{z}_l^K\}_{l = 1}^N$. 

\begin{figure}[!t]
  \centering
  \begin{subfigure}{0.38\textwidth}
    \centering
    \includegraphics[width=\linewidth]{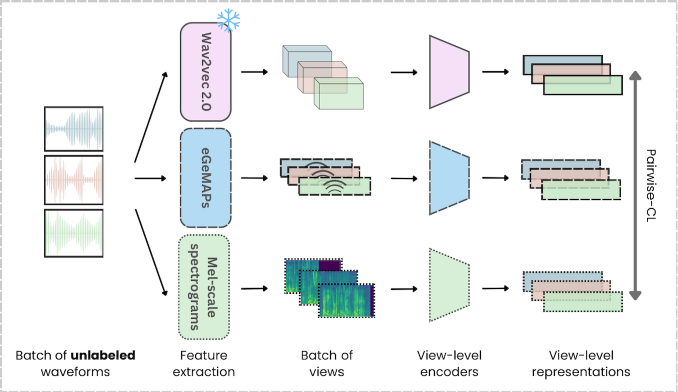}
    \caption{Multi-view pre-training with Pairwise-CL with the selected views: wav2vec 2.0, eGeMAPS-88, and mel spectrograms.}
    \label{fig:pre-training}
  \end{subfigure}
  
  \begin{subfigure}{0.38\textwidth}
    \centering
    \includegraphics[width=\linewidth]{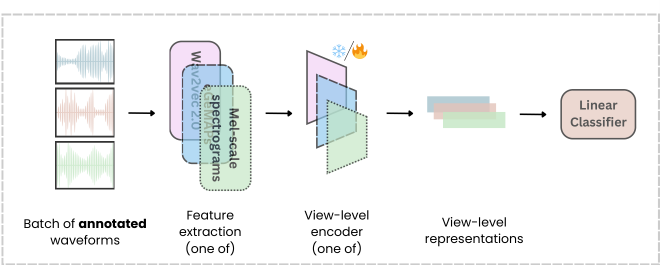}
    \caption{Fine-tuning or supervised training for one of the views. The view-level encoders can be either frozen or fine-tuned with a classifier.}
    \label{fig:ft}
  \end{subfigure}
  \caption{The proposed multi-view SSL framework for speech emotion recognition.}
  \label{fig:framework}
\end{figure}

A pair of projected representations $\boldsymbol{z}_l^i$ and $\boldsymbol{z}_l^j$ is considered positive as they correspond to views of the same $l$-th instance in a mini-batch. The NT-Xent loss $l_l^{i \rightarrow j}$ treating $i$-th view from $l$-th example as an anchor can be computed as follows:

\begin{equation}
   l_l^{i \rightarrow j} = - log \frac{\delta({\boldsymbol{z}}_l^{i},{\boldsymbol{z}}_l^{j})}{\sum_{k=1}^N \delta({\boldsymbol{z}}_l^{i},{\boldsymbol{z}}_k^{j})},
   \label{eq:nt-xent}
\end{equation}
where $\delta({\boldsymbol{z}}_l^{i},{\boldsymbol{z}}_l^{j}) = exp(\frac{s({\boldsymbol{z}}_l^{i},{\boldsymbol{z}}_l^{j})}{\tau})$ and $s(\cdot)$ is the cosine similarity function \cite{chen2020simple}. Therefore, the total loss aggregated for the whole mini-batch of views $i$ and $j$ can be averaged as:

\begin{equation}
    L^{i,j} = \frac{1}{N}\sum_{l=1}^N (l_l^{i \rightarrow j} + l_l^{j \rightarrow i})
\end{equation}

Furthermore, each instance $l$ in a mini-batch is represented by $K$ different views. In the proposed Pairwise-CL, we compute losses between all pairs of views, and average them:

\begin{equation}
    \mathcal{L} = \frac{1}{C(K, 2)} \sum_{k=1}^K \sum_{k'=1}^K \mathbb{I}_{k \neq k'} L^{k, k'},
\label{eq:pairwise}
\end{equation}
where $C(K, 2)$ is a number of possible pairs from $K$ views. Therefore, the proposed loss function aims to maximize the similarities for multi-view representations $\{\boldsymbol{z}_l^1, \boldsymbol{z}_l^2, \dots, \boldsymbol{z}_l^K\}$ corresponding to the same $l$-th instance.

\subsection{Utilizing the Proposed Framework}
\label{sec:util
}
The proposed pre-training framework can be applied to any number of speech views. In this study, we evaluate the framework using a combination of three views, namely wav2vec 2.0 features, eGeMAPS-88 low-level descriptors, and mel-scale spectrograms, as shown in Figure \ref{fig:framework}. This choice is based on their ability to capture different characteristics of speech \cite{li2023exploration}.

\begin{figure}
    \centering
    \scalebox{0.55}{
    \includegraphics{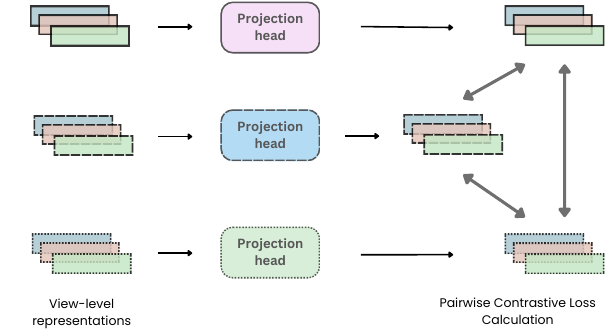}}
    \caption{Pairwise contrastive loss calculation. Representations from each view are first passed through a separate projection head. Later, the contrastive loss is computed in a pairwise fashion, according to Equations \ref{eq:nt-xent} - \ref{eq:pairwise}.}
    \label{fig:pairwise_loss}
\end{figure}

\noindent\textbf{Pre-training.} Representations from each view are processed by a view-specific projection network before computing a pairwise contrastive loss as shown in Figure \ref{fig:pairwise_loss}. Thus, the encoders are trained on unlabeled audio signals to align representations of views from corresponding instances by maximizing cosine similarities among them. We highlight that our approach is only focused on pre-training the view-level encoders, also referred to as downstream architectures for large speech models \cite{zaiem2023speech}. Thus, during pre-training, the wav2vec 2.0 model is frozen and used as a feature extraction method, unlike in relevant studies exploring tuning wav2vec 2.0 parameters \cite{chen2023exploring}. 

\noindent\textbf{Fine-tuning.} Each of the view-level encoders can be fine-tuned by adding a classifier on top of the learnt representations (Figure \ref{fig:ft}). During fine-tuning, the view-level encoders can be either frozen or further tuned via backpropagation using a supervision signal from labeled speech instances.


\section{Implementation Details}
\subsection{Data}
The experiments in this study are based on the IEMOCAP dataset \cite{busso2008iemocap} frequently exploited in the SER literature. The data is collected with 10 subjects in 5 sessions (2 subjects per session). In particular, we use two versions of the dataset in this paper. First, the full version of the dataset, which we refer to as IEMOCAP-10, contains about 10,000 audio samples with 10 distinct emotion annotations. In recent research studies \cite{pepino21_interspeech, morais2022speech} a subset of this dataset with 5,531 samples\footnote{In our experiments, we excluded scripted dialogue \texttt{Ses05M-script01-1} from session 5 containing another trial of the same script as \texttt{Ses05F-script01-1} and \texttt{Ses05M-script01-1b}. All baselines and proposed approaches are evaluated on this subset of data.} and 4 emotions (\textit{neutral}, \textit{angry}, \textit{sad}, and \textit{happy} merged with \textit{excited}), which we denote as IEMOCAP-4, is commonly used. We use IEMOCAP-10 without labels for pre-training purposes, whereas IEMOCAP-4 is mainly used for fine-tuning. We exploit a leave-one-session-out cross-validation (5-fold) protocol consistent between pre-training and fine-tuning data in order to prevent data leakage. In each cross-validation iteration, one session is used for testing purposes, another session is used for validation and early stopping, and three remaining sessions are used for training. We use Unweighted Average Recall (UAR) and Weighted Accuracy (WA) as metrics for SER.

\begin{table*}[!h]
  \begin{minipage}[b]{0.55\textwidth}
    \centering
    \scalebox{0.85}{
    \begin{tabular}{cc|cc|cc|cc}
                             &             & \multicolumn{2}{c|}{wav2vec 2.0} & \multicolumn{2}{c|}{Spectral} & \multicolumn{2}{c}{eGeMAPS-88}  \\
Method                       & $\tau$ & UAR             & WA       & UAR               & WA         & UAR            & WA       \\ \hline
Supervised                   & -           & 62.11           & 59.43          & \textbf{53.39}    & \textbf{52.09}   & 49.97          & 48.1           \\ \hline
\multirow{4}{*}{Pairwise-CL} & 0.1         & 62.24           & \underline{61.51}    & 51.2              & 49.94            & \underline{51.38}    & \textbf{49.69} \\
                             & 0.25        & 62.38           & \textbf{61.59} & 50.69             & 50.17            & 50.77          & 47.99          \\
                             & 0.5         & \textbf{63.11}  & 61.36          & \underline{53.37}       & \underline{51.97}      & \textbf{52.96} & \underline{49.16}    \\
                             & 1.0         & 62.35           & 60.88          & 51.97             & 49.61            & 50.34          & 48.43         
\end{tabular}}
    \caption{Performance metrics for models pre-trained with different values of temperature $\tau$ and fine-tuned on IEMOCAP-4.}
    \label{tab:temp}
  \end{minipage}
  \hfill
  \begin{minipage}[b]{0.4\textwidth}
    \centering
    \scalebox{0.85}{
    \begin{tabular}{cccc}
View                              & \multicolumn{1}{c}{Frozen} & UAR   & WA \\ \hline
\multirow{2}{*}{wav2vec 2.0}      &  \xmark                          & 63.11 & 61.36    \\
                                  &  \cmark                          & 55.43 & 53.97    \\ \hline
\multirow{2}{*}{Spectral} &  \xmark                          & 53.37 & 51.97    \\
                                  &  \cmark                         & 50.34 & 48.43    \\ \hline
\multirow{2}{*}{eGeMAPS-88}       &  \xmark                         & 52.96 & 49.16    \\
                                  &  \cmark                          & 49.65 & 48.97   
\end{tabular}}
    \caption{Results for pre-trained models with frozen and tuned view-level encoders.}
    \label{tab:frozen}
  \end{minipage}
\end{table*}

\subsection{Views and Feature Encoders}
We applied our multi-view pre-training strategy to three views of audio signals downsampled to 16,000 Hz. Each of these views is processed with a view-specific backbone architecture.

\noindent\textbf{wav2vec 2.0.} We use a base version of the wav2vec 2.0 model \cite{baevski2020wav2vec} pre-trained on the LibriSpeech dataset \cite{panayotov_2015_librispeech} available in torchaudio\footnote{\texttt{https://pytorch.org/audio/stable/pipelines.html}}. We trim or pad all audio inputs to a 15-second length \cite{pepino21_interspeech} before feeding them to the model. The generated features are passed through the view-level encoder as proposed in \cite{pepino21_interspeech}. In particular, the outputs of the CNN and transformer blocks are averaged with learnable weights and passed through a two-layer pointwise 1D-CNN, that outputs vectors of size 128.

\noindent\textbf{Mel-scale spectrograms.} The mel spectrograms are extracted with a 25-millisecond window length and a 10-millisecond hop \cite{saeed2021contrastive,fayek2017evaluating}. We employed 64 mel filterbanks covering frequencies from 60 Hz to 7800 Hz. We trim or pad all audio inputs to a 15-second length before generating spectrograms. The obtained spectrograms are then fed to a CNN backbone with three layers.

\noindent\textbf{eGeMAPS-88.} The extended Geneva Minimalistic Acoustic Parameter Set (eGeMAPS-88) \cite{eyben2015geneva} contains 88 derived parameters related to frequency (pitch, jitter), energy, and spectrum aggregated for the whole utterance. We generated these features using the opensmile\footnote{\texttt{https://audeering.github.io/opensmile-python/}} package. A two-layered MLP with 256 and 128 neurons is used as a view-level encoder.

\subsection{Pre-training and Fine-tuning}
During pre-training, we used an MLP with 2 layers of size 256 and 128 as a projection head. The models are pre-trained for 100 epochs with a per-view batch size of 128 and early stopping after 30 epochs with no improvement in validation loss. In the fine-tuning stage, the projection head is dropped and the features are directly passed to a linear classification head with softmax activation. For both pre-training and fine-tuning, we use the Adam optimization algorithm with an initial learning rate of 0.001. During fine-tuning, we decrease the learning rate by a factor of 0.9 after every 5 epochs with no improvement in validation UAR. Pre-training and fine-tuning have been conducted using Nvidia Quadro RTX 5000 GPU (16GB VRAM) with features extracted in advance. With this setup, Pairwise-CL pre-training takes approximately 13 minutes per epoch, whereas fine-tuning time varies based on the used view: 2 seconds for eGEMAPS, 10 seconds for spectrograms, and about 3 minutes for wav2vec 2.0 representations.

\section{Evaluations}

\subsection{Fully Annotated Dataset and Temperature}
\noindent\textbf{Grid search for temperature.} First, we conduct experiments to identify the optimal value of temperature $\tau$ in the contrastive loss function (Equation \ref{eq:nt-xent}). We pre-trained view-level encoders on IEMOCAP-10 with $\tau \in \{0.1, 0.25, 0.5, 1.0\}$ and fine-tuned them on IEMOCAP-4 with all available annotations. The measured metric values are outlined in Table \ref{tab:temp}. In this experiment, the parameters of view-level encoders were not frozen. The first row in the table corresponds to the supervised models trained on each view with the same view-level encoder and classifier architectures. As can be seen from the table, the pre-trained models obtain higher performance in terms of UAR and Weighted Accuracy on wav2vec 2.0 and eGeMAPS-88 features. Besides, pre-trained model performance (with temperature $\tau = 0.5$) is comparable when using mel spectrograms. These results demonstrate that the proposed pre-training strategy, in some cases, can further improve performance when large annotated datasets are available for both pre-training and fine-tuning. Besides, the models pre-trained with temperature $\tau = 0.5$ achieve the highest or the second-highest results for almost all views and metrics. Thus, these pre-training settings will be further explored in the subsequent experiments.

\noindent\textbf{Fine-tune or freeze?} To evaluate the feature representations learnt by encoders on the SSL task only, we compare tuned and frozen encoders during fine-tuning in Table \ref{tab:frozen}. As can be seen, the framework with frozen encoders is less effective. In particular, fine-tuning the view-level encoder on top of the wav2vec features leads to the largest improvement (almost 8\% UAR), compared to the frozen view-level encoder. However, it is worth mentioning that the proposed pre-training allows us to obtain about 50-55\% UAR for all views without tuning the encoders with labels. The gap between the models is less notable (about 3\% UAR) for eGeMAPS-88 and mel spectrograms.

\begin{figure}[!t]
  \centering
  \begin{subfigure}{0.18\textwidth}
    \centering
    \includegraphics[width=\linewidth]{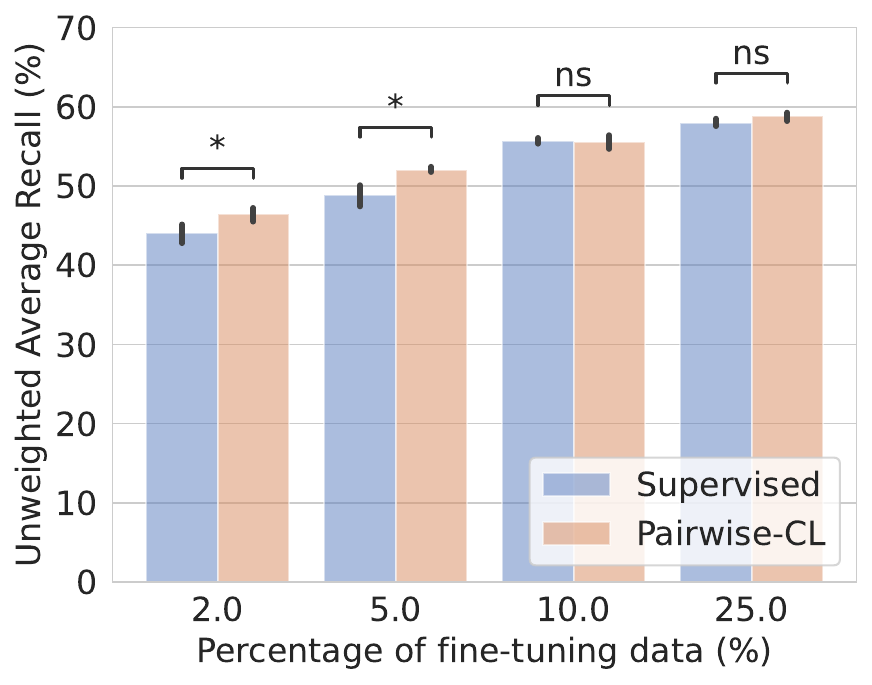}
    \caption{wav2vec 2.0}
    \label{fig:limited_w2v}
  \end{subfigure}
  \begin{subfigure}{0.18\textwidth}
    \centering
    \includegraphics[width=\linewidth]{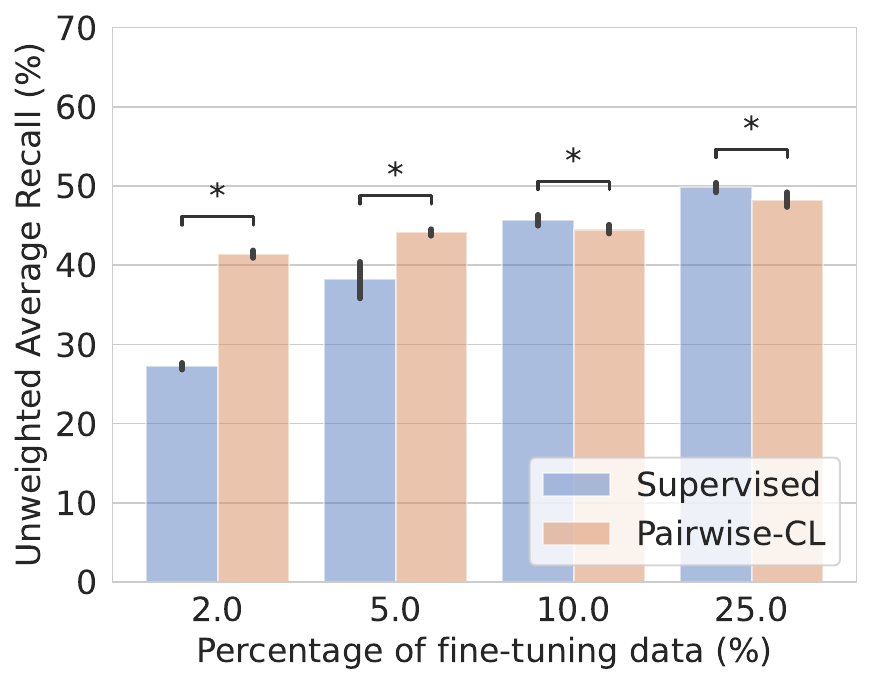}
    \caption{Mel spectrograms}
    \label{fig:limited_spec}
  \end{subfigure}
  \begin{subfigure}{0.18\textwidth}
    \centering
    \includegraphics[width=\linewidth]{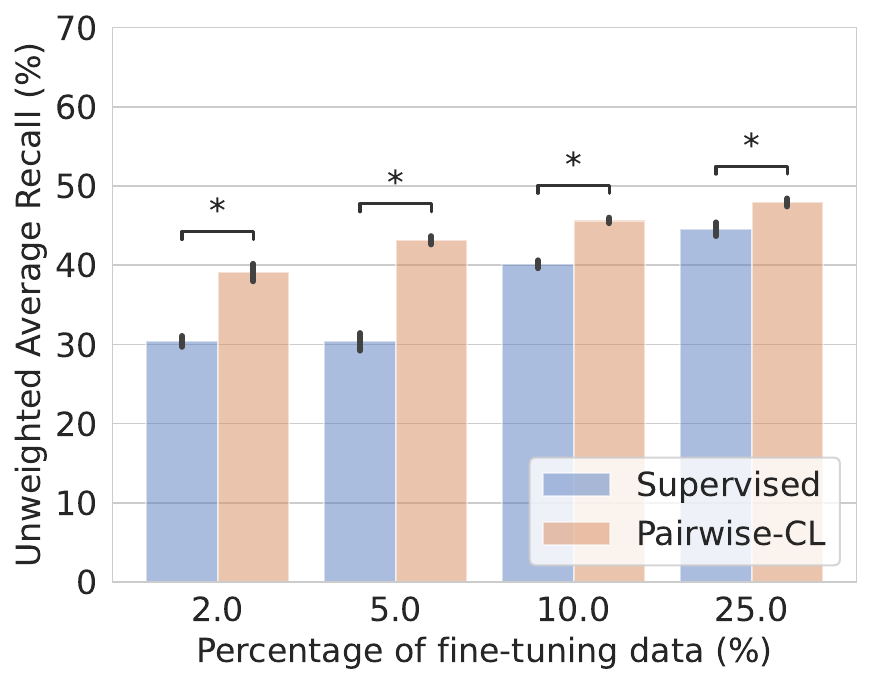}
    \caption{eGeMAPS-88}
    \label{fig:limited_egemaps}
  \end{subfigure}

  \caption{UAR for fine-tuning with limited amounts of labeled data: \textit{*} - statistically significant differences, \textit{ns} - not significant.}
  \label{fig:limited}
\end{figure}

\subsection{Limited Annotated Data}
\label{sec:limited}
\noindent\textbf{Pairwise-CL vs Supervised.} The main motivation of our study is to suggest a pre-training strategy for settings with small amounts of labeled data. Thus, we simulate the scenario with limited annotated data available for fine-tuning by using $p \in \{2\%, 5\%, 10\%, 25\%\}$ of training data from each class in IEMOCAP-4. We fine-tune the pre-trained encoder and train the supervised encoder from scratch models 10 times for each proportion of annotations $p$. In Figure \ref{fig:limited}, we report the average UAR values obtained for each $p$ along with 95\% confidence intervals. Besides, we conduct the Mann-Whitney U-test ($\alpha = 0.05$) to check for statistically significant differences between the supervised and pre-trained models.

According to the obtained metrics, the proposed pre-training strategy significantly improves UAR for all three views in cases with extremely limited annotations ($p \in \{2\%, 5\%\}$). In these cases, the fine-tuning data amounts to approximately 100 and 250 labeled examples per training set (3 session folds). The performance gaps are particularly high for handcrafted features, where improvements reach up to 10-15\% in UAR. For spectral features, supervised models outperform pre-trained ones starting from 10\% of annotations available, whereas, for eGeMAPS, the pre-training strategy is beneficial for all values of $p$.

\begin{figure}[!t]
  \centering
  \begin{subfigure}{0.18\textwidth}
    \centering
    \includegraphics[width=\linewidth]{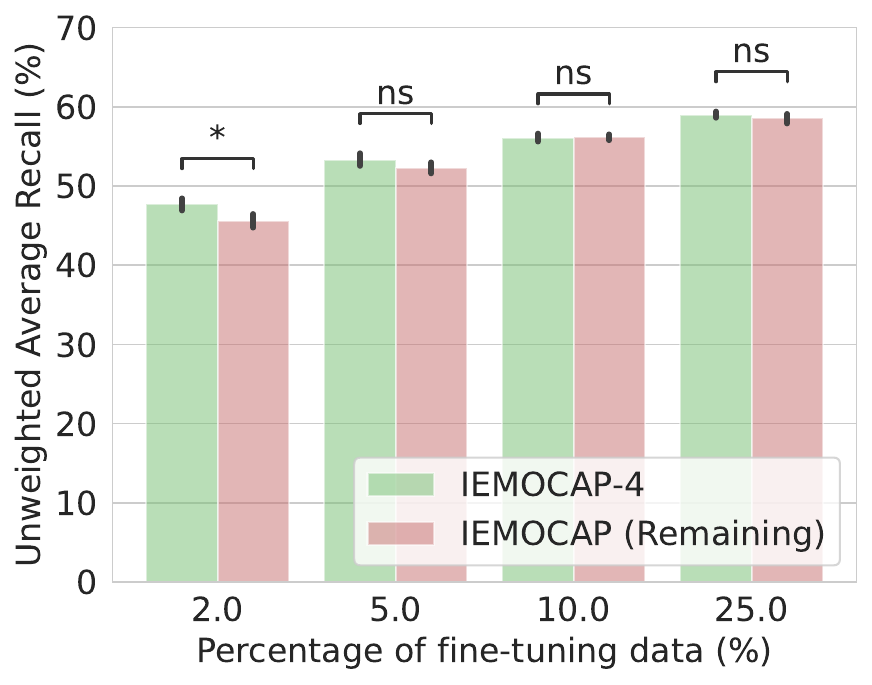}
    \caption{wav2vec 2.0}
    \label{fig:subfiga}
  \end{subfigure}
  \begin{subfigure}{0.18\textwidth}
    \centering
    \includegraphics[width=\linewidth]{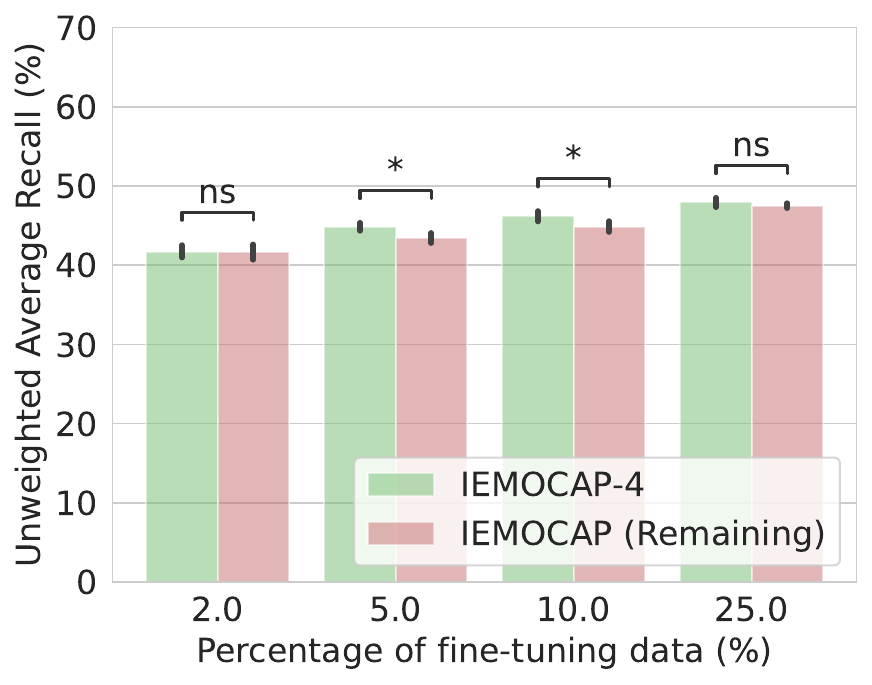}
    \caption{Mel spectrograms}
    \label{fig:subfigb}
  \end{subfigure}
  \begin{subfigure}{0.18\textwidth}
    \centering
    \includegraphics[width=\linewidth]{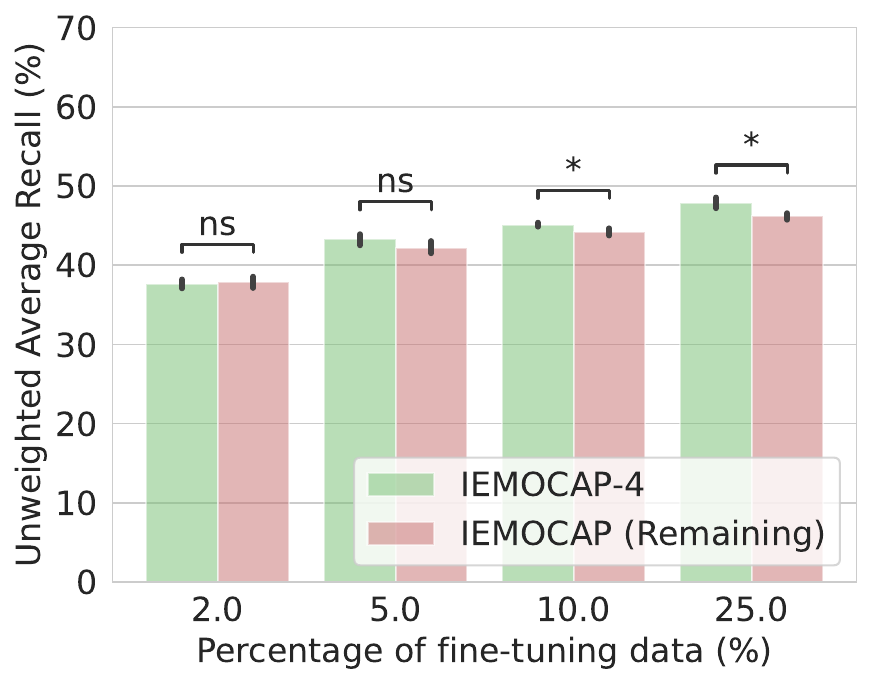}
    \caption{eGeMAPS-88}
    \label{fig:subfigc}
  \end{subfigure}

  \caption{Comparison of model pre-trained on datasets with target (green) and out-of-distribution (red) annotations.}
  \label{fig:limited_dist}
\end{figure}

\noindent\textbf{Pre-training data distribution.} In the previous experiment, the models were pre-trained on IEMOCAP-10 which contains 10 emotions, from which 5 (\textit{happy} and \textit{excited} are merged) are presented in the fine-tuning IEMOCAP-4 dataset. Thus, the remaining emotions are not relevant for fine-tuning. Even though such a scenario represents a realistic case when only some parts of the dataset are annotated, it is interesting to explore how the distribution of pre-training data affects the performance on downstream emotions. In particular, we conduct pre-training on IEMOCAP-4 containing target emotions only. Furthermore, we pre-train another set of encoders on the remaining part of IEMOCAP with out-of-distribution emotions only. We compare both sets of models after fine-tuning them with sparse annotations on IEMOCAP-4 (Figure \ref{fig:limited_dist}). On average, models pre-trained on target distribution data show comparable or better performance, with statistically significant differences observed for wav2vec 2.0 at 2\% of annotations, spectrograms at 5\% and 10\%, and eGeMAPS at 10\% and 25\%. Nevertheless, the gaps in performance for the most sparse annotations are generally smaller compared to the ones reported in Figure \ref{fig:limited}. Thus, model pre-training with target emotions is beneficial but does not lead to large improvements when annotations are limited.

\subsection{View-level Representations and Alignment}

\begin{figure}[!t]
  \centering
  \begin{subfigure}{0.12\textwidth}
    \centering
    \includegraphics[width=\linewidth]{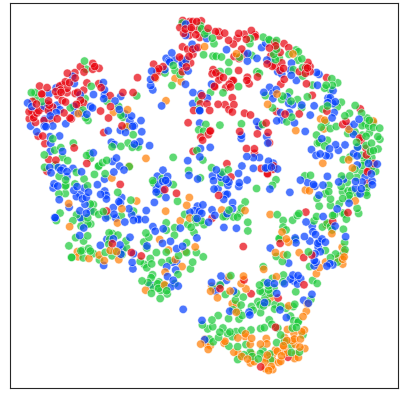}
    \caption{wav2vec 2.0}
    \label{fig:subfiga}
  \end{subfigure}
  \begin{subfigure}{0.12\textwidth}
    \centering
    \includegraphics[width=\linewidth]{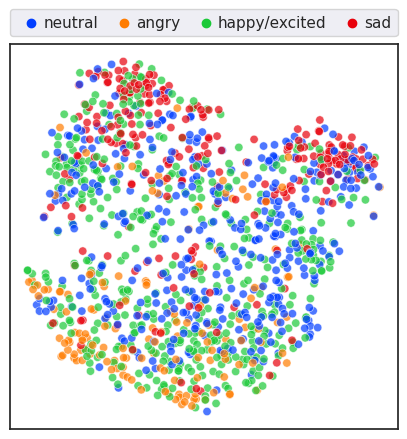}
    \caption{Spectral}
    \label{fig:subfigb}
  \end{subfigure}
  \begin{subfigure}{0.12\textwidth}
    \centering
    \includegraphics[width=\linewidth]{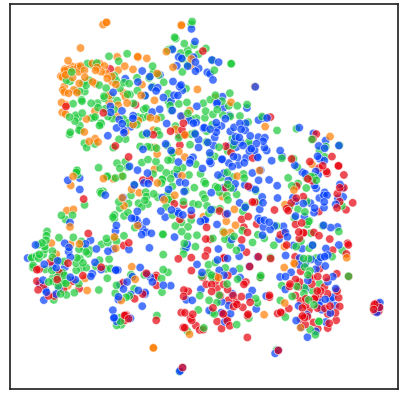}
    \caption{eGeMAPS-88}
    \label{fig:subfigc}
  \end{subfigure}

  \centering
  \begin{subfigure}{0.12\textwidth}
    \centering
    \includegraphics[width=\linewidth]{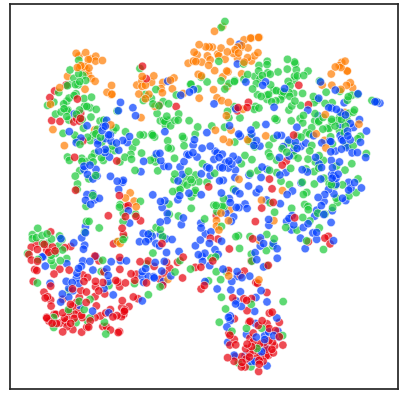}
    \caption{wav2vec 2.0}
    \label{fig:subfiga}
  \end{subfigure}
  \begin{subfigure}{0.12\textwidth}
    \centering
    \includegraphics[width=\linewidth]{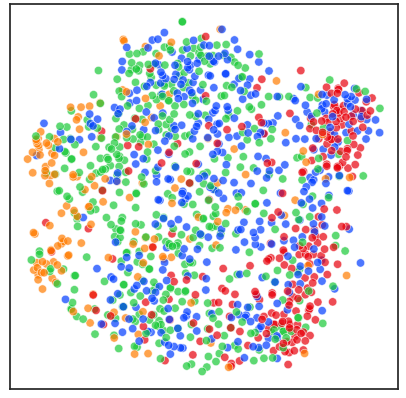}
    \caption{Spectral}
    \label{fig:subfigb}
  \end{subfigure}
  \begin{subfigure}{0.12\textwidth}
    \centering
    \includegraphics[width=\linewidth]{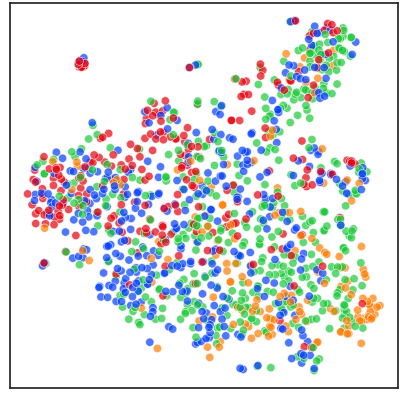}
    \caption{eGeMAPS-88}
    \label{fig:subfigc}
  \end{subfigure}

  \caption{Representations from the test set projected onto the two-dimensional space using t-SNE: (a)-(c) -- Pairwise-CL (before fine-tuning); (d)-(f) -- supervised training from scratch.}
  \label{fig:tsne}
\end{figure}

Figure \ref{fig:tsne} demonstrates the representations learnt by pre-trained and supervised view-level encoders in a two-dimensional space using t-SNE \cite{van2008visualizing}. The supervised models were trained on fully annotated IEMOCAP-4. The illustrated data points correspond to the unseen test subjects from the last cross-validation fold.

The proposed pre-training strategy aims to align representations of different audio signal views. We utilize a projection-weighted Canonical Correlation Analysis (PWCCA) to quantify their alignment. PWCCA has been introduced in \cite{morcos2018insights} as a technique for identifying common structures in features and exploring the similarities between deep representations. Table \ref{tab:pwcca} compares the PWCCA scores obtained for representations of view-level encoders after pre-training to those of supervised models trained independently. The highest level of alignment is obtained for the combination of wav2vec 2.0 and spectral views. Interestingly, even though the PWCCA scores are comparable for pairs with eGeMAPS, there are significant gains in performance for this view after pre-training according to Figure \ref{fig:limited_egemaps}.

\begin{table}[!h]
\centering
\scalebox{0.85}{
\begin{tabular}{cccc}
            & w2v2-egemaps & w2v2-spec & spec-egemaps \\ \hline
Pairwise-CL & 0.652        & 0.782     & 0.640         \\
Supervised  & 0.686        & 0.657     & 0.644        \\
Random      & 0.299        & 0.389     & 0.390         
\end{tabular}}
\caption{PWCCA scores computed for pairs of view-level representations on test instances. The scores are averaged across folds. As a baseline, random scores were computed for pairs of randomly generated vectors of matching shapes.}
\label{tab:pwcca}
\end{table}

\section{Conclusion}
In this paper, we introduced a lightweight contrastive SSL strategy to refine representations of speech in SER settings with sparsely annotated data. Specifically, we evaluated the strategy for three types of views, namely eGeMAPS-88, mel spectrograms, and wav2vec 2.0 features, capturing diverse characteristics of speech. Our experiments demonstrate that the proposed Pairwise-CL technique significantly improves the SER performance when low amounts of annotated data are available. For future work, we suggest experimenting with more views of speech and consider including more modalities during pre-training. 

\newpage

\section{Acknowledgements}
This work has been conducted within the XR2Learn project funded by the European Union’s Horizon Innovation Actions program, under Grant Agreement N.101092851.

\bibliographystyle{IEEEtran}
\bibliography{mybib}

\begin{table*}[t]
\centering
\scalebox{0.84}{
\begin{tabular}{cllcccccccccccc|cc}
\multicolumn{1}{l}{}           &                                     &                                                                            & \multicolumn{4}{c}{wav2vec 2.0}                                                                                                                              & \multicolumn{4}{c}{Spectral}                                                                                                                                  & \multicolumn{4}{c|}{eGeMAPS-88}                                                                                                                               & \multicolumn{2}{c}{Average Ranks}                                           \\
\multicolumn{1}{l}{}           &                                     & \multicolumn{1}{l|}{}                                                      & \multicolumn{2}{c}{Val}                                                      & \multicolumn{2}{c}{Test}                                                      & \multicolumn{2}{c}{Val}                                                       & \multicolumn{2}{c}{Test}                                                      & \multicolumn{2}{c}{Val}                                                       & \multicolumn{2}{c|}{Test}                                                     &                                      &                                      \\
\multicolumn{1}{l}{Method}     & $\tau$                 & \multicolumn{1}{l|}{Frozen}                                                & UAR                                   & WA                                   & UAR                                   & WA                                    & UAR                                   & WA                                    & UAR                                   & WA                                    & UAR                                   & WA                                    & UAR                                   & WA                                    & \multirow{-2}{*}{Val}                & \multirow{-2}{*}{Test}               \\ \hline
\multicolumn{1}{l}{Supervised} & -                                   & \multicolumn{1}{l|}{-}                                                     & 61.35                                 & 60.72                                & \multicolumn{1}{l}{62.11}             & 59.43                                 & 55.11                                 & 54.59                                 & 53.39                                 & 52.09                                 & 48.62                                 & 48.1                                  & 49.97                                 & 48.1                                  & 4.67                                 & 4.17                                 \\ \hline
                               & 0.1                                 & \multicolumn{1}{l|}{\xmark}                                 & 62.49                                 & 62.06                                & 62.24                                 & 61.51                                 & 53.36                                 & 52.69                                 & 51.2                                  & 49.94                                 & 51.32                                 & 51.09                                 & 51.38                                 & 49.69                                 & \underline{2.5}                                  & \underline{2.83}                                 \\
                               &                                     & \multicolumn{1}{l|}{\cmark}                                 & 55.67                                 & 54.93                                & 57.18                                 & 55.13                                 & 47.78                                 & 47.79                                 & 47.45                                 & 47.27                                 & 49.08                                 & 48.63                                 & 50.05                                 & 48.76                                 & 6.67                                 & 6                                    \\
                               & 0.25                                & \multicolumn{1}{l|}{\xmark}                                 & 62.88                                 & 62.46                                & 62.38                                 & 61.59                                 & 53.17                                 & 52.64                                 & 50.69                                 & 50.17                                 & 50.85                                 & 49.72                                 & 50.77                                 & 47.99                                 & 3                                    & 3.67                                 \\
                               &                                     & \multicolumn{1}{l|}{\cmark}                                 & 57.11                                 & 56.41                                & 56.77                                 & 54.59                                 & 47.2                                  & 47.81                                 & 47.07                                 & 46.99                                 & 49.54                                 & 49.58                                 & 48.99                                 & 49.02                                 & 6                                    & 6.83                                 \\
                               & {{0.5}} & \multicolumn{1}{l|}{{{\xmark}}} & {{62.73}} & {{61.8}} & {{63.11}} & {{61.36}} & {{54.89}} & {{54.06}} & {{53.37}} & {{51.97}} & {{50.66}} & {{50.13}} & {{52.96}} & {{49.16}} & {\textbf{2.33}} & {\textbf{1.83}} \\
                               &                                     & \multicolumn{1}{l|}{\cmark}                                 & 55.6                                  & 55.05                                & 55.43                                 & 53.97                                 & 47.15                                 & 47.37                                 & 50.34                                 & 48.43                                 & 48.81                                 & 48.47                                 & 49.65                                 & 48.97                                 & 7.67                                 & 6.5                                  \\
                               & 1                                   & \multicolumn{1}{l|}{\xmark}                                 & 62.22                                 & 60.9                                 & 62.35                                 & 60.88                                 & 53.99                                 & 53.07                                 & 51.97                                 & 49.61                                 & 50.6                                  & 49.97                                 & 50.34                                 & 48.43                                 & 3.5                                  & 4.17                                 \\
\multirow{-8}{*}{Pairwise CL}  &                                     & \multicolumn{1}{l|}{\cmark}                                 & 55.4                                  & 54.83                                & 54.33                                 & 53.21                                 & 47.39                                 & 47.33                                 & 47                                    & 46                                    & 47.57                                 & 46.72                                 & 48.74                                 & 45.91                                 & 8.67                                 & 9                                   
\end{tabular}}
\caption{Fine-tuning results for frozen and tuned view-level encoders pre-trained with different temperature values. For each column with UAR and WA, ranks were calculated by sorting the reported scores. In the last two columns, average ranks obtained for validation and test metrics are presented.}
\label{tab:full_supp}
\end{table*}

\section{Supplementary Materials}
\subsection{Number of parameters}
In Table \ref{tab:model_size}, we present the number of frozen and trainable parameters in the utilized models during pre-training and fine-tuning. As highlighted in Section 2.2, we did not tune the parameters of wav2vec 2.0 and used it as a feature encoder. Thus, the number of frozen parameters in all models exploiting this architecture is no less than the number of wav2vec 2.0 parameters (94.5 million). It can be seen, that the proposed pre-training method outperforms plain fine-tuning on limited data (Figure 3a from the paper) by tuning a small number of parameters corresponding to the view-level encoder applied on top of wav2vec 2.0 features.

\begin{table}[h]
\centering
\scalebox{0.77}{
\begin{tabular}{ccccc}
View                          & Encoder                        & \multicolumn{1}{c|}{Frozen}                & \# frozen (M) & \# trainable (M) \\ \hline
\multicolumn{5}{c}{Supervised Training   / Fine-tuning}                                                                                        \\ \hline
\multirow{2}{*}{wav2vec 2.0}  & \multirow{2}{*}{Pointwise CNN} & \multicolumn{1}{c|}{\xmark} & 94.4          & 0.132            \\
                              &                                & \multicolumn{1}{c|}{\cmark} & 94.515        & 0.017            \\
\multirow{2}{*}{Spectrograms} & \multirow{2}{*}{CNN}           & \multicolumn{1}{c|}{\xmark} & 0             & 0.061            \\
                              &                                & \multicolumn{1}{c|}{\cmark} & 0.035         & 0.026            \\
\multirow{2}{*}{eGeMAPS-88}   & \multirow{2}{*}{MLP}           & \multicolumn{1}{c|}{\xmark} & 0             & 0.029            \\
                              &                                & \multicolumn{1}{c|}{\cmark} & 0.011         & 0.018            \\ \hline
\multicolumn{5}{c}{Pre-training}                                                                                                               \\ \hline
Multi-view                    & All encoders                   & \multicolumn{1}{c|}{-}                     & 94.5          & 0.202           
\end{tabular}}
\caption{Model sizes during supervised training, fine-tuning and pre-training with Pairwise CL. The number (\#) of frozen and trainable parameters is presented in millions (M).}
\label{tab:model_size}
\end{table}

\subsection{Fully-annotated Fine-tuning: Extended Results}
In Tables 1 and 2 from the paper, we demonstrated the summary of the fine-tuning results averaged over unseen test folds in leave-one-session-out cross-validation settings of IEMOCAP-4. Specifically, we tried out different temperature values and freezing or tuning the view-level encoders during fine-tuning. In Table \ref{tab:full_supp} (next page), we present a more thorough summary of the results given all possible combinations between these hyperparameters along with the average metrics obtained on validation and test sessions. Furthermore, we computed the performance ranks (1 -- highest metric score, 9 -- lowest) of models for each metric and data split, and averaged them for validation and test. The average ranks are presented in the last two columns of the table. As can be seen, the highest ranks on both validation (2.33) and test (1.83) data correspond to the model pre-trained with temperature $\tau=0.5$, which have been further used in the experiments with sparse annotations (Section 4.2 from the paper).

\subsection{Randomly Initialized Representations}
\begin{figure}[!h]
  \centering
  \begin{subfigure}{0.13\textwidth}
    \centering
    \includegraphics[width=\linewidth]{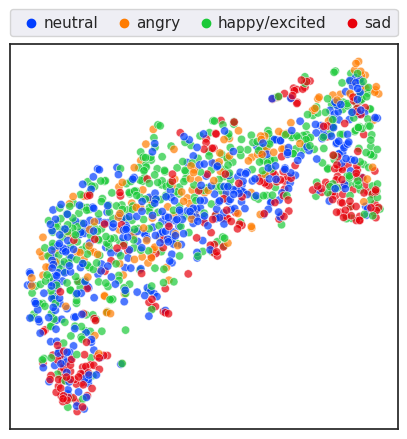}
    \caption{wav2vec 2.0}
    \label{fig:subfiga}
  \end{subfigure}
  \begin{subfigure}{0.13\textwidth}
    \centering
    \includegraphics[width=\linewidth]{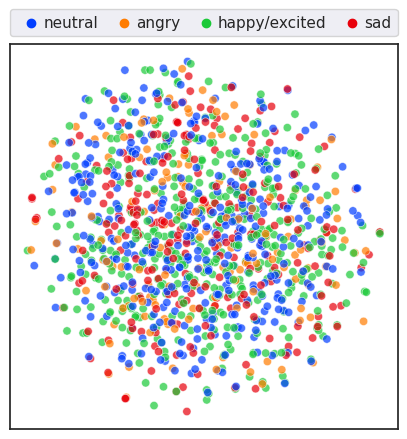}
    \caption{Spectral}
    \label{fig:subfigb}
  \end{subfigure}
  \begin{subfigure}{0.13\textwidth}
    \centering
    \includegraphics[width=\linewidth]{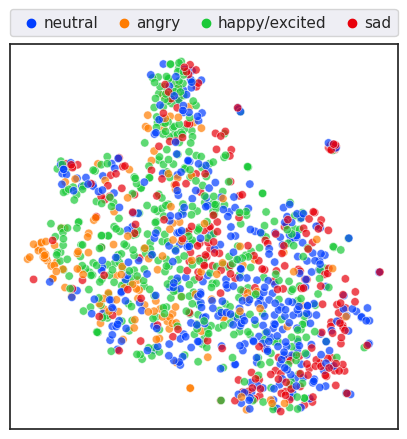}
    \caption{eGeMAPS-88}
    \label{fig:subfigc}
  \end{subfigure}
  \caption{Representations of randomly initialized view-level encoders from the test set projected onto 2D-space using t-SNE.}
  \label{fig:tsne_rand}
\end{figure}

In Figure 5 from the paper, we visualize the feature representations produced by the view-level encoder after pre-training and supervised learning from scratch. As a baseline, in Figure \ref{fig:tsne_rand}, we also demonstrate the t-SNE scatter plot right after random initialization of vier-level encoders, i.e. before any type of training has been applied to them. According to the figure, eGeMAPS-88 and wav2vec 2.0 representations have some initial structure. For wav2vec 2.0, this can be explained by the fact that this method has already been pre-trained on raw speech, whereas eGeMAPS-88 is a set of features that extract handcrafted features meaningful for recognizing emotions. In the case of the spectrograms, the initial representations do not reflect any patterns. Nevertheless, the proposed SSL pre-training strategy contributes to better grouping of representations bringing them closer to what can be achieved with fine-tuning or supervised training on the fully annotated dataset.

\end{document}